# LLM attribution analysis across different fine-tuning strategies and model scales for automated code compliance

Jack Wei Lun Shi[1], Minghao Dang[1,2], Wawan Solihin[1,3], Justin K.W. Yeoh[1]

[1] Department of Civil and Environmental Engineering, National University of Singapore, 117576, Singapore
[2] School of Transportation Science and Engineering, Harbin Institute of Technology, Harbin 150090, China
[3] Research and Innovation, NovaCITYNETS Pte.Ltd., 348615, Singapore
jackswl@u.nus.edu

**Abstract**

Existing research on large language models (LLMs) for automated code compliance has primarily focused on performance, treating the models as black boxes and overlooking how training decisions affect their interpretive behavior. This paper addresses this gap by employing a perturbation-based attribution analysis to compare the interpretive behaviors of LLMs across different fine-tuning strategies such as full fine-tuning (FFT), low-rank adaptation (LoRA) and quantized LoRA fine-tuning, as well as the impact of model scales which include varying LLM parameter sizes. Our results show that FFT produces attribution patterns that are statistically different and more focused than those from parameter-efficient fine-tuning methods. Furthermore, we found that as model scale increases, LLMs develop specific interpretive strategies such as prioritizing numerical constraints and rule identifiers in the building text, albeit with performance gains in semantic similarity of the generated and reference computer-processable rules plateauing for models larger than 7B. This paper provides crucial insights into the explainability of these models, taking a step toward building more transparent LLMs for critical, regulation-based tasks in the Architecture, Engineering, and Construction industry.

## 1    Introduction

The emergence of large language models (LLMs) has introduced a new approach for automated code compliance (ACC), offering promising solutions to the conventional labor-intensive process of manual rule interpretation. Recent studies have demonstrated the potential and versatility of LLMs to convert complex building rules into computer-processable rules. These applications include identifying predefined atomic functions to interpret rules [1], a prompt-based framework to transform rules into Prolog [2], directly generating scripts from rules via fine-tuning an LLM on domain-specific rule-script pairs [3], and using a multi-modal LLM to derive computer-processable rules from a combination of textual and graphical information [4].

Despite these remarkable advancements, existing research has focused solely on evaluating the performance and accuracy of the generated outputs, often treating the LLMs as black boxes. It remains unclear how different training decisions such as the choice between full fine-tuning versus parameter-efficient fine-tuning (PEFT) methods or the effect of model scale (e.g., 3 billion vs. 22 billion parameters) influence the internal interpretative mechanisms of the LLM. This knowledge gap limits our ability to understand and trust these systems, which is a significant barrier to their adoption for critical, regulation-bound tasks in the Architecture, Engineering, and Construction (AEC) industry.

To address this knowledge gap, this paper utilizes a perturbation-based attribution analysis to compare the interpretive behaviors of LLMs across different fine-tuning strategies and model scales in the context of rule interpretation. We found that the attribution patterns of fully fine-tuned models are statistically different from those using PEFT methods. Additionally, our results indicate that model scale dictates which parts of the building regulatory text are prioritized, with larger models exhibiting distinct attribution patterns compared to smaller models.

## 2    Literature Review

Existing research in ACC leverages LLMs through several approaches to automate the traditionally manual process of rule interpretation. For example, Lee and Lee [5] addressed context loss in retrieval-augmented generation (RAG) systems by employing long-context models to process entire legal documents, thereby preserving crucial cross-references. Similarly, Kim et al. [4] demonstrated the value of multi-modality by utilizing LLMs to derive computer-processable rules from both textual rules and their accompanying graphical representations, achieving higher accuracy than text-only methods.

To enhance the performance of LLMs on specialized ACC tasks, few research focuses on fine-tuning for specific applications. The primary challenge with this approach is the high computational cost of full fine-tuning (FFT), which requires significant computational resources. Consequently, PEFT methods have become a popular alternative, as they modify only a small subset of the model's parameters, making the fine-tuning process more computationally efficient. For example, Shi et al. [3] utilized quantized low-rank adaptation (QLoRA) fine-tuning to adapt a 7 billion parameter model for converting textual building rules into Python scripts. The QLoRA technique was chosen to overcome the high computational costs associated with the long and complex rule-script pairs characteristic of ACC tasks. In a different context, Lee et al. [6] applied the LoRA technique to fine-tune a 12.8 billion parameter model for retrieving construction safety management knowledge. Their goal was to adapt the model to a specialized question-answer dataset derived from safety guidelines, demonstrating the efficiency of PEFT methods on limited computational resources.

While these studies validate the efficacy of various training strategies based on performance metrics, it remains unclear how different fine-tuning strategies and model scales affect the models' internal interpretive mechanisms. Therefore, this paper addresses this gap by using perturbation-based attribution to identify which specific words of an input building rule are most influential in generating the corresponding computer-processable rule. By systematically analyzing these attribution patterns, we can gain a deeper understanding of what textual features the models prioritize and how this focus changes across different configurations.

This paper investigates the following research questions: (1) How do different fine-tuning strategies and model scales influence the learned attribution patterns of fine-tuned LLMs when converting textual building rules to computer-processable rules? (2) Which specific textual features within building rules do these LLMs prioritize, and how does this interpretative focus shift across different training configurations? By addressing these questions, this paper provides crucial insights into the explainability of LLMs for ACC, moving beyond performance metrics to understand the interpretive mechanisms behind their outputs.

## 3 Methodology

To investigate the interpretive behaviors of LLMs in ACC, the methodology is structured into three stages. This systematic process allows us to first establish model performance and then analyze the underlying attribution patterns. The stages are: (1) preparation of the fine-tuned models and dataset, (2) quantitative performance evaluation based on semantic similarity of generated and reference scripts, and (3) explainability analysis using perturbation-based attribution. Our implementation uses the Captum library, an open-source model explainability tool for PyTorch [7].

### 3.1 Models and Dataset

We prepared a range of LLMs fine-tuned on a domain-specific real-world dataset of building regulations paired with their corresponding Python script implementations, an extension of our previous work [3]. The models were configured to test two key variables: fine-tuning strategy (i.e., FFT, LoRA fine-tuning, and QLoRA fine-tuning), and model scale (i.e., varying LLM parameter counts). The summary of the experimental configurations is shown in Table 1.

**Table 1.** LLM configurations for experimental analyses. The attribution analysis compared fine-tuning strategies on only the LLaMA 3B model and model scales across the three designated models.

| Parameter Size | Instructional LLM Families | Fine-tuning Strategies Comparison (FFT, LoRA, QLoRA) | Model Scale Comparison (QLoRA) | Model Used for Attribution Analysis |
|---|---|---|---|---|
| < 3B | Qwen (0.5B, 1.5B) | – | ✓ | – |
| 3B | Falcon, Hermes, LLaMA, Phi, Qwen | ✓ | ✓ | LLaMA 3B |
| 7B – 8B | LLaMA, Mistral, Qwen | – | ✓ | Mistral 7B |
| > 10B | Nemo-Mistral (12B), Mistral-Small (22B) | – | ✓ | Mistral-Small 22B |

Our experimental design involved two primary comparisons. To evaluate model scale, we applied QLoRA fine-tuning across ten models ranging from 0.5B to 22B parameters: Qwen (0.5B, 1.5B, 3B, 7B), LLaMA (3B, 8B), Hermes (3B),

Mistral (7B), Nemo-Mistral (12B), and Mistral-Small (22B). To compare fine-tuning strategies, we applied FFT, LoRA, and QLoRA across five 3B parameter models: LLaMA, Hermes, Falcon, Phi, and Qwen. For the computationally intensive attribution analysis, we selected a representative subset, performing the model scale analysis on LLaMA 3B, Mistral 7B, and Mistral-Small 22B, while focusing the fine-tuning strategy analysis exclusively on the LLaMA 3B variants. All models were fine-tuned on a real-world dataset of rule-script pairs, each consisting of a textual building rule (i.e., input) and its corresponding Python script (i.e., output).

### 3.2 Performance Evaluation

The primary evaluation of the generated code's quality is performed using CodeBERTScore [8]. This metric leverages contextual embeddings from the CodeBERT model to measure semantic similarity between the generated and reference scripts. Unlike lexical token-matching metrics, CodeBERTScore is better correlated with functional correctness and human preference [8], making it a robust choice for evaluating code generation tasks. We conducted this performance analysis across all models under investigation.

### 3.3 Perturbation-based Attribution via Feature Ablation

To quantitatively analyze the models' interpretive behaviors, we employ a perturbation-based attribution method which assesses the contribution of each word within an input building rule to the generation of the target Python script. This approach does not require access to model gradients and instead relies on observing the model's output behavior in response to controlled input perturbations, hence speeding up computations.

The attribution algorithm involves systematically isolating each word in the building rule, treating it as a distinct feature, and measuring its influence on the model's output. The model's output is quantified as the log probability of generating the entire target script sequence. The influence of a single word is determined by ablating it from the input and measuring the resulting change in the log probability of the target script. To focus the attribution analysis on meaningful terms, common stop words (e.g., 'the') were excluded from the ablation process.

Take a fine-tuned LLM that computes the conditional probability of a target sequence $T$ (i.e., the Python script) given an input sequence $X$ (i.e., the building rule), where $X$ consists of $d$ words $\{w_1, w_2, ..., w_d\}$. Let $f(X)$ represent the log probability of the target, such that $f(X) = logP(T \mid X)$. The attribution score $\phi_i$ for the $i$-th word $w_i$ is calculated as the difference in this log probability when the word is present versus when it is ablated:

$$\phi_i(f, X) = f(X) - f(X_{D\setminus\{i\}})$$

$$= logP(T \mid X) - logP(T \mid X_{D\setminus\{i\}}) \quad (1)$$

In Equation 1, $D$ represents the set of all word indices $\{1, 2, ..., d\}$, and $X_{D\setminus\{i\}}$ thus denotes the input sequence $X$ with the $i$-th word removed, resulting in a shorter sequence for that evaluation step. A positive attribution score $\phi_i$ indicates that the word $w_i$ increased the model's confidence in the generation of $T$, thus signifying its importance as a positive contributor. Conversely, a negative attribution score $\phi_i$ suggests that the word's presence hindered the generation of $T$. Scores near zero imply the word had negligible effect.

This procedure yields a set of attribution scores, one for each word in the input regulation. The resulting attribution vector provides a quantitative and interpretable link between specific textual features of a building rule and the model's final Python output, enabling a granular inspection of what the model has learned to prioritize during inferencing.

## 4 Results and Discussion

### 4.1 Semantic Analysis

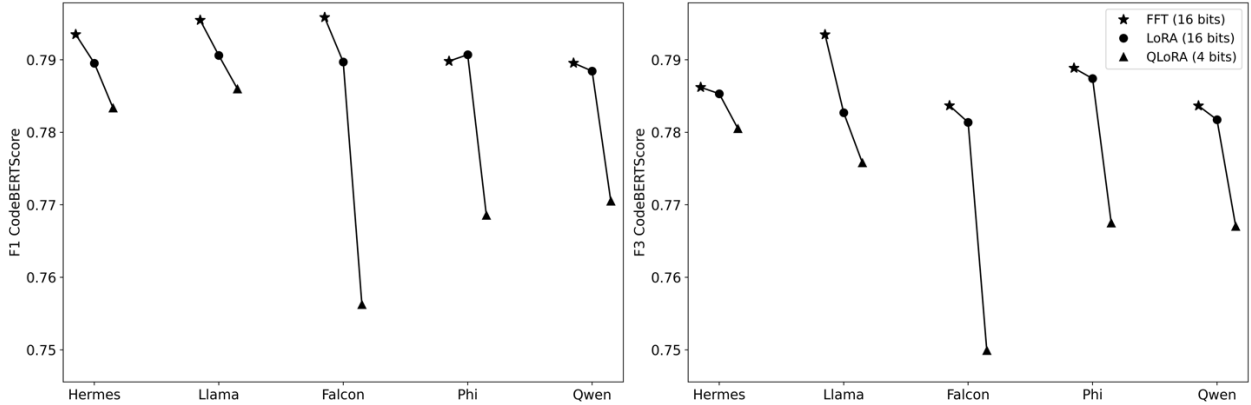

**Fig. 1.** Comparison for different fine-tuning strategies based on semantic similarity of generated script and reference script.

The semantic similarity between the generated and reference scripts, as measured by CodeBERTScore F1 and F3, reveals a clear performance hierarchy across the fine-tuning strategies. As shown in Fig. 1, FFT consistently achieved the highest scores (except for Phi under F1 score), indicating that updating the full model parameters allows for the most accurate semantic mapping from building regulation text to computer-processable rules. This is followed by LoRA, which, while parameter-efficient, shows a slight degradation in performance. QLoRA consistently performed the worst, suggesting that the 4-bit quantization introduces a significant loss of fidelity.

The performance gap between LoRA and QLoRA is not uniform across models as it is notably more pronounced for Falcon, Phi, and Qwen. This further suggests that certain model architectures may be more sensitive to the precision loss inherent in quantization. For a complex task such as rule interpretation, the information lost during quantization appears to have a negative impact on the quality of the generated code.

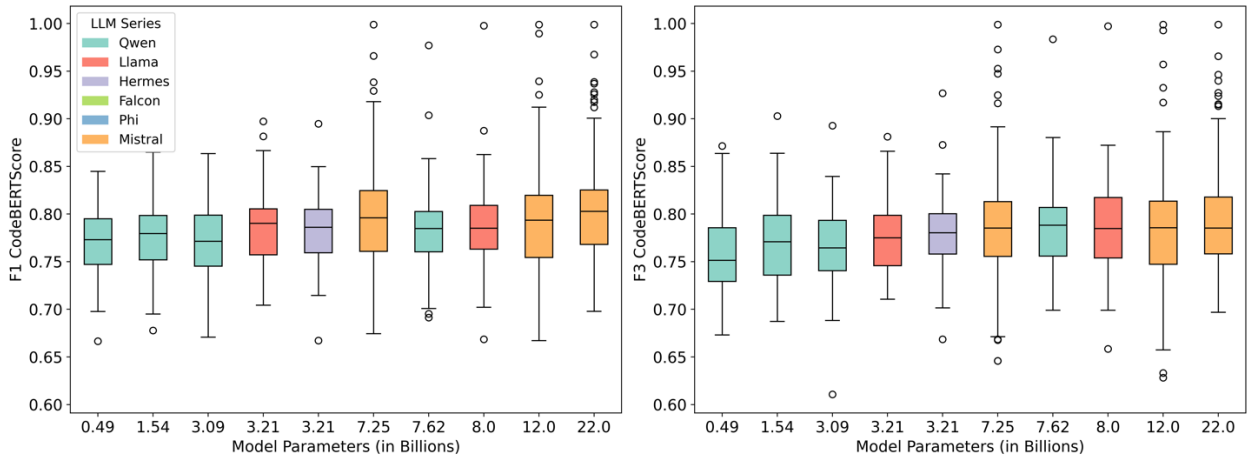

**Fig. 2.** Comparison across varying parameter sizes for LLMs based on semantic similarity of generated script and reference script.

Subsequently, our investigation into model scaling, conducted only using QLoRA fine-tuning, reveals a nuanced relationship between parameter count and semantic similarity, as shown in Fig. 2. A slight positive correlation between model size and performance is evident in the sub-7B parameter range. Median F1 and F3 CodeBERTScore demonstrate an upward trend from the 0.5B model to the ~7B models. However, this scaling advantage appears to diminish beyond this point, with performance plateauing for models between 7B and 22B.

This plateau suggests a point of diminishing returns for this specific task when constrained by QLoRA. While the initial performance increase aligns with the principle that larger models have a greater capacity for complex reasoning, the subsequent stagnation may indicate several possibilities. The complexity of the task itself might be adequately addressed by a ~7B parameter model, with larger models offering no significant additional benefit. It could also be possible that the intricacy of the task exceeded a threshold where merely scaling model size is effective in itself, given that the dataset is derived a real-world implementation and contains inherent variations in coding style and logic.

Alternatively, the performance ceiling could be an artifact of the QLoRA method itself, where the 4-bit quantization and limited scope of trainable parameters may prevent the larger models from fully leveraging their architectural depth. Furthermore, the notable increase in high-performing outliers for the 7B, 12B, and 22B models is generally more pronounced compared to the sub-7B models. This suggests that while the median performance does not improve, these larger models possess a higher potential to generate exceptionally accurate scripts for specific cases, even if this capability does not translate much to an overall increase in average performance.

### 4.2   Attribution Analysis

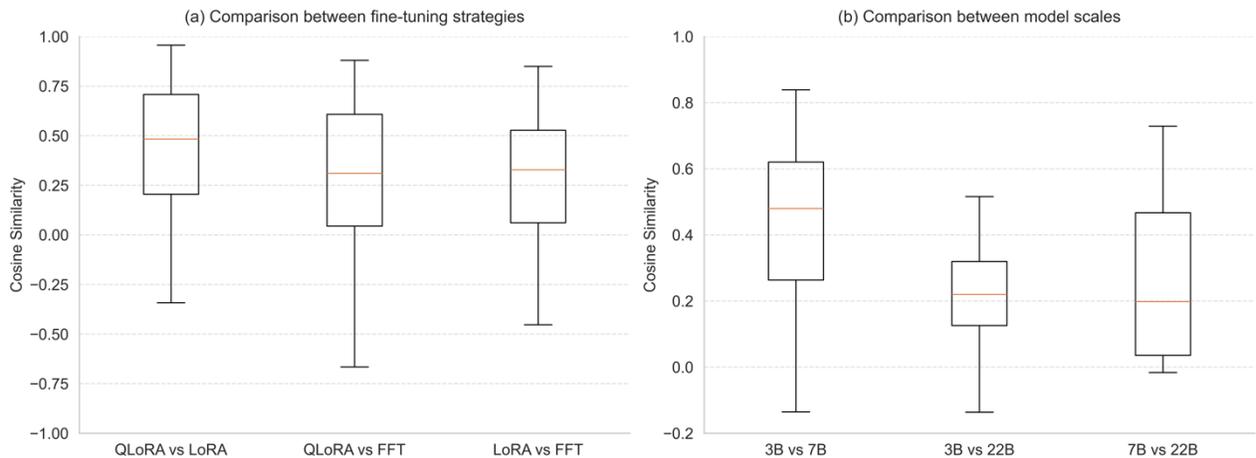

**Fig. 3.** a) Comparison for cosine similarity of the attribution vectors between the fine-tuning strategies across all 79 test samples and b) comparison for cosine similarity of the attribution vectors across the three model scales, however only for 15 test samples.

We now quantitatively investigate the attribution analysis. Fig. 3 shows the box plots for comparison between QLoRA vs LoRA, QLoRA vs FFT, and LoRA vs FFT. A high positive cosine similarity means despite different fine-tuning strategies, the models can converge on an almost identical attribution pattern, focusing on the same words with similar relative importance within the building rule. In contrast, a high negative cosine similarity means the fine-tuning strategies lead to opposite attribution patterns. In other words, the words that one fine-tuning strategy found highly influential (i.e., large positive score) are found highly suppressive (i.e., large negative score) in another. A cosine similarity close to 0 indicates uncorrelated attribution, suggesting that the models are focusing on entirely different and unrelated sets of input words to inform their output, indicating no systematic relationship between their learned influence patterns. The average cosine similarity is 0.451, 0.288, 0.293 respectively, showing that generally, QLoRA and LoRA have relatively similar attribution patterns compared to FFT. Statistical testing via paired t-tests and Wilcoxon signed-rank tests show that the similarity between QLoRA and LoRA is significantly different from their respective similarities to FFT (p-value < 0.001), while the difference between QLoRA vs FFT and LoRA vs FFT comparison is not statistically significant (p-value > 0.50).

Next, we observe the attribution patterns qualitatively as shown in Fig. 4 to Fig. 6. For each sample, we generated a visualization where each input word is underlined with a color intensity proportional to the magnitude of its attribution score. To ensure a fair comparison across the fine-tuning strategies, the color scale is normalized per sample. Specifically, for a given sample, we first identify the maximum absolute attribution score across all three models (i.e., FFT, LoRA, and QLoRA). This single maximum value is then used to set the endpoint of the color scale (i.e., white for a score of zero, red for the maximum score) for all three visualizations within that sample. This approach allows for a direct visual comparison of attribution magnitudes across models, revealing how differently each fine-tuning strategy directs the model's prioritization of textual features on the same input text.

```
FFT
SSW 4.3.4 (d) (iv) No discharge pipe connection vertical bends: 4.3.4 Discharge Stacks Offsetsd. Horizontal
offset, any, comply following requirements:iv. No discharge pipe connected discharge stack 500mm vertical
bends horizontal offsite pipe 2.5m vertical bends.

LoRA
SSW 4.3.4 (d) (iv) No discharge pipe connection vertical bends: 4.3.4 Discharge Stacks Offsetsd. Horizontal
offset, any, comply following requirements:iv. No discharge pipe connected discharge stack 500mm vertical
bends horizontal offsite pipe 2.5m vertical bends.

QLoRA
SSW 4.3.4 (d) (iv) No discharge pipe connection vertical bends: 4.3.4 Discharge Stacks Offsetsd. Horizontal
offset, any, comply following requirements:iv. No discharge pipe connected discharge stack 500mm vertical
bends horizontal offsite pipe 2.5m vertical bends.
```

**Fig. 4.** An example of a building rule pertaining to discharge stacks.

```
FFT
SWD 10.1 (a) Pumping capacity: 10 Pumped Drainage System 10.1 minimum design operation criteria pumped
drainage system follows: (a) pumping capacity adequate cater immediate discharge storm water ingress not less
than 150 millimetres per hour entire source catchment area; i.e.: P __IA___ 3.6 x 106 P pumping capacity (m3/
s) I rainfall intensity (mm/hr) catchment area contributing ingress storm water (m2)

LoRA
SWD 10.1 (a) Pumping capacity: 10 Pumped Drainage System 10.1 minimum design operation criteria pumped
drainage system follows: (a) pumping capacity adequate cater immediate discharge storm water ingress not less
than 150 millimetres per hour entire source catchment area; i.e.: P __IA___ 3.6 x 106 P pumping capacity (m3/
s) I rainfall intensity (mm/hr) catchment area contributing ingress storm water (m2)

QLoRA
SWD 10.1 (a) Pumping capacity: 10 Pumped Drainage System 10.1 minimum design operation criteria pumped
drainage system follows: (a) pumping capacity adequate cater immediate discharge storm water ingress not less
than 150 millimetres per hour entire source catchment area; i.e.: P __IA___ 3.6 x 106 P pumping capacity (m3/
s) I rainfall intensity (mm/hr) catchment area contributing ingress storm water (m2)
```

**Fig. 5.** An example of a building rule pertaining to pumped drainage system.

```
FFT
7.30.4 Shadow areas Sloping ground: Shadow areas existing undulating sloping terrain sloping ground below
building structures, platform deck excluded gross floor area computation. qualify exemption, exempted shadow
areas cannot enclosed sides. See Figure 7-31 below.

LoRA
7.30.4 Shadow areas Sloping ground: Shadow areas existing undulating sloping terrain sloping ground below
building structures, platform deck excluded gross floor area computation. qualify exemption, exempted shadow
areas cannot enclosed sides. See Figure 7-31 below.

QLoRA
7.30.4 Shadow areas Sloping ground: Shadow areas existing undulating sloping terrain sloping ground below
building structures, platform deck excluded gross floor area computation. qualify exemption, exempted shadow
areas cannot enclosed sides. See Figure 7-31 below.
```

**Fig. 6.** An example of a building rule pertaining to shadow areas.

Fig. 4 illustrates a case of high attributional agreement, where all strategies identify similar key phrases such as 'vertical bends', 'No discharge', and '500mm vertical bends' as highly influential. However, the magnitude of these attributions varies. FFT and LoRA assign higher importance scores to these words compared to QLoRA, suggesting a more decisive feature prioritization. In Fig. 5, the FFT model concentrates its attribution sharply on a few critical terms, namely 'Pumping capacity' and 'rainfall intensity (mm/hr)'. Conversely, LoRA and QLoRA exhibit a more diffuse attribution pattern, distributing importance more evenly across a wider range of words. This widespread attribution may potentially indicate model uncertainty, where no single set of words is deemed significantly more important than others in generating the output. This suggests that models fine-tuned with LoRA and QLoRA may not have learned as strong a latent mapping as the FFT model. This trend is further corroborated by Fig. 6. Here, the FFT model again demonstrates a concentrated attribution strategy, focusing primarily on 'Shadow areas', 'Sloping ground', and 'cannot enclosed sides'. This pattern, where a few words contribute most to the output, indicates that FFT has learned to rely on a concise set of the most salient textual features compared to LoRA and QLoRA.

7We now compare the cosine similarity for model scale for 3B LLM, 7B LLM, and 22B LLM. Due to computational constraints, generating the attribution for large models require significant amount of time, therefore in Fig. 3b, only 15 samples were selected. With that, the average cosine similarity for 3B vs 7B, 3B vs 22B, and 7B vs 22B is 0.355, 0.156, and 0.208 respectively. This indicates that there is a much similar attribution pattern for 3B and 7B LLM, whereas for 22B LLM, the attribution pattern is relatively different. This is more pronounced when comparing 3B against 22B.

Next, we observe the attribution patterns across different model scales qualitatively as shown in Fig. 7 to Fig. 9. However, in this case, since models of different scales may produce attribution scores of different magnitudes, a shared color scale would obscure the patterns of certain models. Therefore, to facilitate a fair comparison of their internal attribution patterns, the color scale is normalized individually for each model. In other words, for any given sample, the visualization for the 3B model is colored based on its own maximum attribution score, and likewise for 7B model and 22B model. This approach allows us to clearly compare the relative importance each model assigns to the input words, revealing how a model's attribution patterns evolve as its scale increases, independent of the absolute score magnitudes.

```
3B LLM
SSW 4.3.11d Floor trap on open areas 4.3.11 Floor Trap Shallow Floor Trap and Floor Waste d The floor trap
shall not be located in an open area receiving rainwater or surface runoffs

7B LLM
SSW 4.3.11d Floor trap on open areas 4.3.11 Floor Trap Shallow Floor Trap and Floor Waste d The floor trap
shall not be located in an open area receiving rainwater or surface runoffs

22B LLM
SSW 4.3.11d Floor trap on open areas 4.3.11 Floor Trap Shallow Floor Trap and Floor Waste d The floor trap
shall not be located in an open area receiving rainwater or surface runoffs
```

**Fig. 7.** An example of a building rule pertaining to floor trap.

```
3B LLM
BCA C.3.2.2 Minimum headroom of parking lots and driveway for car parks For sheltered car parks the headroom
at parking lots and driveway shall not be less than 2.2 metres

7B LLM
BCA C.3.2.2 Minimum headroom of parking lots and driveway for car parks For sheltered car parks the headroom
at parking lots and driveway shall not be less than 2.2 metres

22B LLM
BCA C.3.2.2 Minimum headroom of parking lots and driveway for car parks For sheltered car parks the headroom
at parking lots and driveway shall not be less than 2.2 metres
```

**Fig. 8.** An example of a building rule pertaining to headroom of parking lots and driveway for car parks.

```
3B LLM
SCDF 4.2.2 a (3) (b) Fire engine accessway length and distance from the facade Clause 4.2.2 (a)(3) PG II
buildings exceeding 10m habitable height For a building under PG II that exceeds the habitable height of 10m
all of the following shall be complied with (b) A fire engine accessway of at least ¼ length of perimeter
minimum 15m whichever is greater shall be provided to access at least one façade of each block and shall be
located at a distance of at least 2m and at most 10m away from the façade of the building This is to
facilitate rescue with direct access to unit windows excluding exit staircase smoke-free approach to exit
staircase

7B LLM
SCDF 4.2.2 a (3) (b) Fire engine accessway length and distance from the facade Clause 4.2.2 (a)(3) PG II
buildings exceeding 10m habitable height For a building under PG II that exceeds the habitable height of 10m
all of the following shall be complied with (b) A fire engine accessway of at least ¼ length of perimeter
minimum 15m whichever is greater shall be provided to access at least one façade of each block and shall be
located at a distance of at least 2m and at most 10m away from the façade of the building This is to
facilitate rescue with direct access to unit windows excluding exit staircase smoke-free approach to exit
staircase

22B LLM
SCDF 4.2.2 a (3) (b) Fire engine accessway length and distance from the facade Clause 4.2.2 (a)(3) PG II
buildings exceeding 10m habitable height For a building under PG II that exceeds the habitable height of 10m
all of the following shall be complied with (b) A fire engine accessway of at least ¼ length of perimeter
minimum 15m whichever is greater shall be provided to access at least one façade of each block and shall be
located at a distance of at least 2m and at most 10m away from the façade of the building This is to
facilitate rescue with direct access to unit windows excluding exit staircase smoke-free approach to exit
staircase
```

**Fig. 9.** An example of a building rule pertaining to fire engine accessway length and distance from the facade.

ICCCBE, 23-26 March 2026, Taipei, Taiwan                                                                                133-7

Fig. 7 highlights a consistent understanding of the rule's semantics across different model scales, while also revealing a more abstract strategy in the largest model. Both the 3B and 7B models identify key components of the rule, assigning high attribution to the negative constraint 'not' and critical phrases like 'surface runoffs' and 'open area'. In addition to this shared understanding, the 22B model is unique in that it also assigns high importance to the rule identifier '4.3.11d'. This possibly suggests the model has learned better association, recognizing that the rule identifier is itself a powerful and predictive feature.

In Fig. 8, we also observe that all models demonstrate a consensus on the core concepts of the rule but differ in their ability to pinpoint critical parameters. All three models seem to attribute high importance to the primary subjects: 'headroom', 'sheltered car parks', and the rule identifier 'C.3.2.2'. This indicates that the fundamental topic of the rule is easily identified regardless of model scale. The 22B model, however, is unique in its strong attribution to the numerical constraint '2.2 metres'. While smaller models understand the concept of headroom, the 22B model has learned that the specific value is the most critical word for generating a relevant script. This demonstrates a shift from general conceptual understanding to relevant parameter extraction, possibly a more advanced capability crucial for rule interpretation.

In Fig 9, we observe a trend of attribution sharpening with increased model scale. While the 3B model's attribution is broadly distributed across many words, the 7B and 22B models exhibit a more focused pattern with fewer, but stronger, attribution peaks. This widespread attribution in 3B model may suggest model uncertainty or a less refined understanding, where it cannot decisively identify the most critical words. In contrast, the 7B and 22B models exhibit relatively highly concentrated attribution. In both, they identify 'perimeter' as the single most influential word and assign it a significantly higher score than other words. This may indicate a more confident and efficient interpretive strategy, where the models have successfully learned to isolate the core feature that dictates the logic of the entire script.

## 5    Conclusion

This paper investigates LLM attribution for ACC, revealing that both fine-tuning strategy and model scale alter a model's interpretive behavior. We found that FFT results in attribution patterns that are statistically distinct from PEFT methods, and that larger models may learn more sophisticated strategies compared to smaller models. It is imperative to highlight, however, that these attribution scores are not a measure of correctness. Rather, they provide insight into the model's internal focus. Further research is needed to determine if these findings generalize to other ACC domains while exploring the practical balance between a model's interpretive depth and the high computational resource demands of FFT and increased model scale. Nevertheless, the findings in this paper represent a step toward building more transparent and explainable LLMs for critical tasks in the AEC industry.

## References

1. Zheng, Z., Han, J., Chen, K.Y., Cao, X.Y., Lu, X.Z. and Lin, J.R.: Translating regulatory clauses into executable codes for building design checking via large language model driven function matching and composing. *Engineering Applications of Artificial Intelligence 163* (2026).
2. Yang, F., and Jiansong Z.: Prompt-based automation of building code information transformation for compliance checking. *Automation in Construction* 168 (2024).
3. Shi, J.W.L., Solihin, W. and Yeoh, J.K.W.: Fine-tuning a large language model for automated code compliance of building regulations. *Advanced Engineering Informatics* 68 (2025).
4. Kim, Y., Borrmann, A. and Lee, G.: A preliminary study on design rule derivation from graphical representations using multimodal large language models. Proceedings of the 2025 European Conference on Computing in Construction, Porto, Portugal (2025).
5. Lee, J., and Ghang L.: Long context window-based zero-shot legal interpretation of building codes and regulations. *Automation in Construction* 179 (2025).
6. Lee, J., Ahn, S., Kim, D. and Kim, D.: Performance comparison of retrieval-augmented generation and fine-tuned large language models for construction safety management knowledge retrieval. *Automation in Construction* 168 (2024).
7. Miglani, V., Yang, A., Markosyan, A.H., Garcia-Olano, D. and Kokhlikyan, N.: Using captum to explain generative language models. *arXiv preprint arXiv:2312.05491* (2023).
8. Zhou, S., Alon, U., Agarwal, S. and Neubig, G.: Codebertscore: evaluating code generation with pretrained models of code. *Proceedings of the 2023 Conference on Empirical Methods in Natural Language Processing* (2023).